\documentclass[runningheads]{llncs}
\usepackage{amsmath}
\usepackage{amssymb}
\usepackage{mathrsfs}
\usepackage[T1]{fontenc}
\usepackage{bbold}
\usepackage[all]{xy}
\usepackage{graphicx}
\numberwithin{equation}{section}
\numberwithin{figure}{section}
\begin{document}
\title{Linguistic Loops and Geometric Invariants as a Way to Pre-Verbal Thought?}
\titlerunning{Linguistic Loops for Pre-Verbal Thought}
%
\author{Daniele Corradetti\inst{1,2,3} \and Alessio Marrani\inst{1,4} }
\authorrunning{D. Corradetti et al.}
%
\institute{Elementar, Divisione Ricerca e Sviluppo,  10121 Torino, Italy
\and Departamento de Matematica, Universidade do Algarve, Campus de Gambelas, 8005-139 Faro, Portugal
\and Grupo de Física Matemática, Instituto Superior Técnico, Av. Rovisco Pais, 1049-001 Lisboa, Portugal
\and Dipartimento di Management 'Valter Cantino', Università degli Studi di Torino, Corso Unione Sovietica 218 bis, 10134 Torino }
\maketitle              

\begin{abstract}
We introduce the concepts of linguistic transformation, linguistic loop and
semantic deficit. By exploiting Lie group theoretical and geometric
techniques, we define invariants that capture the structural properties of a
whole linguistic loop. This result paves the way to a totally new line of
research, employing tools from Lie theory and higher-dimensional geometry
within language studies. But, even more intriguingly, our study hints to a
mathematical characterization of the meta-linguistic or pre-verbal thought,
namely of those cognitive structures that precede the language.
\end{abstract}

\section{Introduction}

When Raymond Queneau published his \textquotedblleft \emph{Exercices de style%
}\textquotedblright\ in 1947, inspired by Bach's \textquotedblleft \emph{Art
of Fugue}\textquotedblright , he had the intuition to be doing something
intrinsically geometric (so much, that he named the first version of his
work \textquotedblleft \emph{Dod\'{e}ca\`{e}dre}\textquotedblright ). This
intuition, far from being casual, may be regarded as the first example of
what in this study we define as \emph{linguistic transformation}. Similarly
to a geometric transformation, which transforms an object while maintaining
some of its invariant algebraic and geometric properties, a linguistic
transformation is, in essence, a map (usually expressed through a
proposition in the natural language) that, when applied to an element of a
linguistic space, transforms it into another, while preserving in a
controlled way the original semantic core, through criteria of coherence and
reversibility. Therefore, linguistic transformations are transformations of
propositions or text elements characterized by a limited generative content,
since they mainly act on the form or signifier of the sentence, with a
controlled (and generally limited) impact on its meaning. Moreover, the
aforementioned requirement of coherence and reversibility of such
transformations yields to the introduction of the concept of \emph{%
linguistic loop}, associated to geometric invariants which can be explicitly
(and effectively) calculated. As such, the linguistic loop displays strong
similarities to the Wilson loop, introduced by the physicist K. G. Wilson in
1974 in his lattice computations in Quantum Field Theory \cite{Wilson}, and
subsequently vastly employed in Differential Geometry, Mathematical Physics
\cite{Rudolph}, and also in lattice formulations of nonperturbative Quantum
Gravity \cite{Modanese,Ambjorn,Hamber}.

From a theoretical point of view, the use of concepts and tools from Lie
theory and higher-dimensional geometry, usually employed in the formulation
of (classical and quantum) theories of gravity (even beyond Einstein's
General Relativity), represents an interesting and new perspective within
the study of linguistic spaces. More than this, tantalizing hints arise to
an even more intriguing development, namely to provide a formal, rigorous
starting point for the study of cognitive structures pertaining to the \emph{%
pre-verbal} thought, which transcends any formulation through language.
Indeed, given a loop of linguistic transformations that change completely
the form of a sentence while leaving its meaning unaltered, one might ask
whether it is possible to characterize the \emph{semantic core} of the
sentence by means of some invariant structure(s), preserved by the loop
itself.

While it is reasonable to state that the existence of two systems of
reasoning, namely system 1 (`fast') and system 2 (`slow'), is gaining a
quite broad recognition within the scientific communty (see e.g. \cite%
{Kahneman}), we cannot help but observe that the same community is
essentially unaware of another equally, if not more, important type of
thought system, namely the \emph{meta-linguistic}, or (as we prefer to name
it) \emph{pre-verbal}, \emph{thought}.

For our purposes, pre-verbal thought refers to mental representations and
cognitive structures that exist before being expressed in words. A careful
analysis of the mental process giving rise, for example, to the formulation
of an elaborate speech, immediately yields one to realize that the speech,
before being lexically structured, consists of a \emph{pre-verbal} intention
that only eventually is clothed in a specific language (according to the
grammar and the vocabulary learnt by the speaker). In this way, a polyglot
can choose whether to formulate his idea in an Italian, Portuguese, English
or French without virtually altering the content and the structure of the
idea he wants to convey. If rational thought takes place on the linguistic
plane in a way made of logical deductions and syllogisms, meta-rational or
abstract or pre-verbal thought takes place independently of any specific
language.

From an experimental point of view, cognitive science and developmental
psychology provide clear evidence that children and even animals form rich
concepts without language \cite{Mandler,Lakusta,Spelke,Piaget,Vygotsky}.
This phenomenon finds its confirmation in neuroscience applied to the
brain-computer interfacer, when it uses semantic vector spaces as
intermediaries by decoding brain activity into text. In the research
conducted in \cite{Fesce,Tang}, for example, cerebral magnetic resonance
models (non-verbal signals) are mapped into a semantic integration that is
then translated by a linguistic model into sentences that describe the
person's thoughts. These `\emph{proto-concepts}', identified in neuroscience
as a pattern of neural activation indicating a `face'{} or a `hand'{} before
any language is employed, can be modeled in AI by nodes or embeddings that
respond to certain abstract categories. These proto-concepts have a clear
pre-verbal nature, and they an support rudimentary thought and even
non-verbal communication (e.g., emotional expressions).

At this point, one may wonder whether it makes sense to talk about
pre-verbal thought in the context of linguistic spaces such as those of a
Large Language Model (LLM). Is it possible to identify concepts or invariant
structures that are meta-linguistic, and which therefore pertain to the
pre-verbal thought, within a space defined in terms pure linguistic
elements? In this article, we suggest that this is indeed possible, by
defining suitable invariants of linguistic spaces.

It should be remarked that our use of algebraic-geometric invariants in
linguistic spaces is distinctly different from the one made within usual
studies of differential geometry on higher-dimensional varieties originated
from LLMs. Usually, these studies apply differential geometry techniques to
the embedding space, and they consider some classes of geometric
transformations of the vectors thereof. However, geometric transformations
in the linguistic space do not necessarily correspond to well-defined
linguistic transformations which preserve (or alter in a controlled way) the
semantic core of the linguistic element they act upon. In other words,
purely geometric transformations acting onto embedding vectors are not an
efficient and purposeful tool in the identification/determination of
invariants allowing to identify the pre-verbal meaning of a sentence.

Instead, for a given sentence and a given sequence of re-formulations of
this sentence through the application of a certain set of linguistic
transformations (such as translations, paraphrases, negations, etc.), we are
going to identify some invariants, which thus capture intrinsic properties,
transcending the applied linguistic transformations.

More concretely, it is here worth recalling that in a LLM an \emph{embedding
system} consists of an application from a set of linguistic elements or
dictionary to a vector space \cite{Mikolov,Pennington}. Therefore, through
the embedding map one can switch from a discrete space, i.e., the
dictionary, to a continuous space. It is important to note that the
embedding map is generally \emph{not surjective}, so an element of the
continuous embedding space is not always uniquely translatable into one or
more propositions. This aspect is crucial, as our hypothesis is that the
study of specific linguistic loops (such as, for example, multilingual
translation) can not only allow the identification of geometric invariants
(as we present in this work), but will also allow the identification of an
element in the embedding space that, even though it cannot be formulated
linguistically, can `generate' the whole loop itself (this will be the aim
of our future research).

\section{Linguistic Transformations}

A \emph{metrizable linguistic space} $\left( \mathscr{A},\psi ,d\right) $ is
a linguistic space $\mathscr{A}$, endowed with an embedding map $\psi $ from
$\mathscr{A}$ to $\mathbb{R}^{n}$, along with a metric $\mu $ and a distance
$d$, both defined on $\mathbb{R}^{n}$. The linguistic space $\mathscr{A}$ is
a place where natural language lives, and it is constituted by a set of
linguistic elements (words, phrases, propositions, etc.) in a given context
of interest; for example, $\mathscr{A}_{LLM_{1}}$ can indicate the
linguistic space containing all the propositions that can be generated by a
certain model $LLM_{1}$, while the linguistic space $\mathscr{A}_{\text{Eng}}
$ can indicate the set of all linguistic elements of the English language.
On the other hand, the embedding map $\psi $ is a mathematical function that
maps elements of the linguistic space $\mathscr{A}$ (i.e., elements of
natural language such as words, phrases, or entire propositions) from a
discrete space (such as a dictionary) to a continuous vector space, such as $%
\mathbb{R}^{n}$ :%
\begin{equation}
\psi :\left\{
\begin{array}{l}
\mathscr{A}\rightarrow \mathbb{R}^{n}; \\
\lambda \mapsto \psi \left( \lambda \right) .%
\end{array}%
\right.
\end{equation}%
Note that, generally, $\psi $ is not injective nor surjective.

In this framework, by picking a distance $d_{\ast }$ in $\mathbb{R}^{n}$,
one can define a \emph{semantic distance} $d$ in the linguistic space $%
\mathscr{A}$, simply by composing $\psi $ and $d_{\ast }$ itself :%
\begin{equation}
d\left( \lambda ,\nu \right) :=d_{\ast }\left( \psi \left( \lambda \right)
,\psi \left( \nu \right) \right) ,
\end{equation}%
\ for any $\lambda ,\nu $ in $\mathscr{A}$. For instance, a typical semantic
distance in $\mathscr{A}$ is the one induced by the so-called \emph{cosine
distance} in $\mathbb{R}^{n}$,
\begin{equation}
d\left( \lambda ,\nu \right) :=d_{\ast \text{,cosine}}\left( \psi \left(
\lambda \right) ,\psi \left( \nu \right) \right) =1-\frac{\psi \left(
\lambda \right) \cdot \psi \left( \nu \right) }{\Vert \psi \left( \nu
\right) \Vert \Vert \psi \left( \nu \right) \Vert }.  \label{dist}
\end{equation}

Let us now introduce the concept of \emph{linguistic transformation}. A
linguistic transformation, denoted by $U$, is a transformation of the
signifier of an element of $\mathscr{A}$ that preserves in some way the
semantic core of the element itself. As such, we will see below that a
linguistic transformation is the building block of a \emph{linguistic loop},
and it will therefore be instrumental in the definition of the \emph{%
semantic deficit} of the loop itself. For all this to be possible, we
require a linguistic transformation to enjoy the following features :

\begin{itemize}
\item \emph{Closure} : the image of a linguistic transformation lies in $%
\mathscr{A}$ :%
\begin{equation}
U\left( \lambda \right) \in \mathscr{A},~\forall \lambda \in \mathscr{A},
\end{equation}

\item \emph{Reversibility} : the generative character of a linguistic
transformation is controlled, in such a way that the content and the meaning
of the original element of $\mathscr{A}$ is traceable in some way. More
precisely, for any $U\in \mathbb{U}\left( \mathscr{A}\right) $ there exists
an `inverse' linguistic transformation (denoted, with a certain abuse of
language, $U^{-1}\in \mathbb{U}\left( \mathscr{A}\right) $), for which the
semantic distance between the original linguistic element (say, $\lambda $)
and the correspondingly saturated element $U^{-1}\left( U\left( \lambda
\right) \right) $ is less than a certain defined \emph{threshold }(or, in a
physicist's jargon, `\emph{ultraviolet cutoff}') $\varepsilon \in \mathbb{R}%
^{+}$ :%
\begin{equation}
U\in \mathbb{U}\left( \mathscr{A}\right) ~\text{is~\emph{reversible}}\overset%
{\text{def.}}{\Leftrightarrow }~\exists U^{-1}:d\left( U^{-1}\left( U\left(
\lambda \right) \right) ,\lambda \right) <\varepsilon ,~\forall \lambda \in %
\mathscr{A}.  \label{reversible}
\end{equation}%
This request is necessary to allow for the possibility to return, after the
composition of a certain number of linguistic transformations, to the
original element, or to another element \emph{similar} to it (see next
condition); as we will see below, the reversibility of the linguistic
transformations ultimately allows the linguistic loops to exist. Note that
generally $U^{-1}$ is not uniquely defined, and it also depends on the
choice of $\varepsilon $. In this framework, the concept of `reversible'
transformation is an \emph{approximate} (non-unique, and $\varepsilon $%
-dependent) version of the concept of `inverse'.

\item \emph{Coherence} : the application the same transformation to \emph{%
similar} elements of $\mathscr{A}$ yields to \emph{similar} image elements.
More precisely, by defining as reciprocally\emph{\ similar} any two elements
$\lambda ,\nu \in $ $\mathscr{A}$ such that, for a certain fixed $\epsilon
\in \mathbb{R}^{+}$, $d\left( \lambda ,\nu \right) <\epsilon $, the
condition for a certain transformation $U$ to be coherent is the following
one : for any element $\lambda \in \mathscr{A}$, one can define a ($U$%
-dependent) \emph{similar} element $U^{-1}\left( U\left( \lambda \right)
\right) $ (which always exists if $U$ is reversible; in this case, $\epsilon
=\varepsilon $), and then compute the semantic distance of the images of
such two similar elements under $U$ itself. If $U$ is coherent, such a
distance will be less than a certain function\footnote{%
The subscript \textquotedblleft $U$\textquotedblright\ denotes the fact that
$f$ \emph{a priori} depends on $U$.} $f_{U}\left( \varepsilon \right) $ :
\begin{equation}
U\in \mathbb{U}\left( \mathscr{A}\right) ~\text{is~\emph{coherent}}~\overset{%
\text{def.}}{\Leftrightarrow }d\left( U^{-1}\left( U\left( \lambda \right)
\right) ,\lambda \right) <f_{U}(\varepsilon ),~\forall \lambda \in %
\mathscr{A}.  \label{coherent}
\end{equation}
\end{itemize}

The set $\mathbb{U}\left( \mathscr{A}\right) $ of all linguistic
transformations acting on $\mathscr{A}$ (endowed with the map composition $%
\circ $) is an example of \emph{unital} \emph{magma }(see e.g. \cite{Bergman}%
). Additionally (but not necessarily), one may require a linguistic
transformation to be \emph{identifiable} (or \emph{natural}), namely to be
identified by a linguistic element of the same space on which it acts (i.e.,
of $\mathscr{A}$); for instance, $U_{\mu }$ will indicate the natural
linguistic transformation $U\in \mathbb{U}\left( \mathscr{A}\right) $,
identified by the natural phrase or linguistic element $\mu \in \mathscr{A}$.

To be more concrete, let us now proceed to identify some notable examples of
linguistic transformations : multilingual transformations that convert a
text from one language to another (e.g., \emph{\textquotedblleft translate
from English to Italian\textquotedblright }), are linguistic transformations
if applied to an adequate linguistic space (domain); conversion of
propositions in negative form, as well as conversion of propositions in
interrogative form are linguistic transformations, as well. In general, all
linguistic processes with a limited generative content (and mainly focused
on the linguistic form) are linguistic transformations; for example:
dialectal adaptation, colloquial or formal; poetic transposition or
adaptation in a poetic style; technical transposition or conversion of
common language into technical or specialized language/jargon; expansion of
a text and its synthesis; paraphrasing, namely reformulating the same
content with synonyms or similar linguistic structures preserving the same
meaning; emotional amplification through some specific emotional filter;
etc. Conversely, all those processes that have a too ample generative
component which prevents one from reconstructing the original semantic
content, for example \emph{\textquotedblleft invent a
story\textquotedblright }, are \emph{not} linguistic transformations.

\section{Linguistic Loops and Semantic Deficits}

Given a metrizable linguistic space $\left( \mathscr{A},\psi ,d\right) $, a
\emph{sequence} $\mathcal{U}$ of length $L+1$ is the ordered set of $L+1$
linguistic transformations (always including the identity map $\mathbb{I}$
as the first element), denoted as%
\begin{equation}
\mathcal{U}:=\left\{ \mathbb{I},U_{1},...,U_{L}\right\} .  \label{U-call}
\end{equation}%
Starting from an initial linguistic element $\lambda \in \mathscr{A}$ and
applying the iterated composition of such transformations to $\lambda $, one
obtains a \emph{sequence} of length $L+1$ in the linguistic space $%
\mathscr{A}$,
\begin{equation}
\mathcal{U}\left( \lambda \right) :=\left\{ \lambda ,\;U_{1}\left( \lambda
\right) ,\,\left( U_{2}\circ U_{1}\right) \left( \lambda \right) ,...,\left(
U_{L}\circ U_{L-1}\circ \dots \circ U_{2}\circ U_{1}\right) \left( \lambda
\right) \right\} .  \label{ling-seq}
\end{equation}%
Clearly, this sequence always contains the element $\lambda $. By further
applying the embedding map $\psi $ to each element of $\mathcal{U}\left(
\lambda \right) $, one obtains the corresponding \emph{sequence} $\psi
\left( \mathcal{U}\left( \lambda \right) \right) $ of length $L+1$ in $%
\mathbb{R}^{n}$ :%
\begin{equation}
\psi \left( \mathcal{U}\left( \lambda \right) \right) :=\left\{
v,v_{1},v_{2},...,v_{L-1},v_{L}\right\} ,
\end{equation}%
where
\begin{eqnarray}
v &:&=\psi \left( \lambda \right) ,  \notag \\
v_{1} &:&=\psi \left( U_{1}\left( \lambda \right) \right) ,  \notag \\
v_{2} &:&=\psi \left( \left( U_{2}\circ U_{1}\right) \left( \lambda \right)
\right) ,  \notag \\
&&...  \notag \\
v_{L} &:&=\psi \left( \left( U_{L}\circ U_{L-1}\circ ...\circ U_{2}\circ
U_{1}\right) \left( \lambda \right) \right) .  \label{0-n-loop}
\end{eqnarray}

The semantic distance between the first and the last element of the sequence
(\ref{ling-seq}) in $\mathscr{A}$, namely the quantity%
\begin{equation}
\delta _{\mathcal{U}}\left( \lambda \right) :=d\left( \lambda ,\left(
U_{L}\circ U_{L-1}\circ ...\circ U_{2}\circ U_{1}\right) \left( \lambda
\right) \right) =d_{\ast }\left( v,v_{L}\right)  \label{sem-def}
\end{equation}%
is named \emph{semantic deficit} of the element $\lambda \in \mathscr{A}$
under the sequence $\mathcal{U}$ defined by (\ref{U-call}). If, for a given,
fixed threshold (or, again, \textquotedblleft ultraviolet
cutoff\textquotedblright ) $\xi >0$,%
\begin{equation}
\delta _{\mathcal{U}}\left( \lambda \right) <\xi ,~\forall \lambda \in %
\mathscr{A},  \label{loop}
\end{equation}%
then $\mathcal{U}$ is defined as a \emph{linguistic loop} (of length $L+1$).
In practice, a smaller semantic deficit indicates a greater preservation of
the (meaning of the) original (i.e., starting) linguistic element, while
higher deficits (but still under the threshold $\xi $) signal an accumulated
distortion (of the meaning of the original linguistic element) within the
linguistic loop under consideration.

We will employ the notions of semantic deficit and linguistic loop
linguistic in order to study in a quantitative way the stability and the
reliability of the maps of the linguistic transformations. This will work in
a intriguingly analogous way to what happens for the iteration of maps in
dynamic systems, in which small iterative errors/deficits can lead to
significant twists or deviations from the initial path.

\section{From Linguistic Loops to Quadratic Forms, and their Signatures}

The notion of semantic deficit, and its evaluation on elements of the
linguistic space $\mathscr{A}$ for various linguistic loops defined in the
magma $\mathbb{U}\left( \mathscr{A}\right) $ of all linguistic
transformations, is interesting \emph{per se}. However, the purpose of this
work is to determine \emph{invariant} structures which enjoy some degree of
\emph{stability} under the action of $\mathbb{U}\left( \mathscr{A}\right) $.
As mentioned above, it is our hypothesis that these structures may pertain
to an immutable core of pre-verbal thought or meaning.

Since the inhomogeneous Lie group IO$\left( n\right) :=$O$\left( n\right)
\ltimes T^{n}$ has a transitive action on $\mathbb{R}^{n}$ itself, the
sequence $\psi \left( \mathcal{U}\left( \lambda \right) \right) $ (\ref%
{0-n-loop}) of $L+1$ vectors in $\mathbb{R}^{n}$ can be rewritten as%
\begin{eqnarray}
v &:&=\psi \left( \lambda \right) ,  \notag \\
v_{1} &=&\frac{\left\Vert v_{1}\right\Vert }{\left\Vert v\right\Vert }%
\mathcal{R}_{1}v,  \notag \\
v_{2} &=&\frac{\left\Vert v_{2}\right\Vert }{\left\Vert v_{1}\right\Vert }%
\mathcal{R}_{2}v_{1}=\frac{\left\Vert v_{2}\right\Vert }{\left\Vert
v\right\Vert }\mathcal{R}_{2}\mathcal{R}_{1}v,  \notag \\
&&...  \notag \\
v_{L} &=&\frac{\left\Vert v_{L}\right\Vert }{\left\Vert v_{L-1}\right\Vert }%
\mathcal{R}_{L}v_{L-1}=\frac{\left\Vert v_{L}\right\Vert }{\left\Vert
v\right\Vert }\mathcal{R}_{L}\mathcal{R}_{L-1}...\mathcal{R}_{2}\mathcal{R}%
_{1}v,  \label{n-loop}
\end{eqnarray}%
where $\left\Vert \cdot \right\Vert $ denotes the Euclidean norm in $\mathbb{%
R}^{n}$, and $\mathcal{R}_{i}$ ($i=1,...,L$) is an element of O$\left(
n\right) $. For $n\geqslant 3$, the rotation matrix $\mathcal{R}_{i}$ is not
unique, and so is the rewriting (\ref{n-loop}) of (\ref{0-n-loop}); however,
for any given pair of vectors $x$ and $y$ in $\mathbb{R}^{n}$, there exists
a \emph{unique} definition of \emph{`minimal'}\ rotation matrix $\mathcal{%
\tilde{R}}^{x,y}$ connecting $y$ to $x$, namely
\begin{equation}
y=:\frac{\left\Vert y\right\Vert }{\left\Vert x\right\Vert }\mathcal{\tilde{R%
}}^{x,y}x,
\end{equation}%
where

\begin{equation}
\mathcal{\tilde{R}}^{x,y}:=I_{N}+\Bigl(\hat{y}\,\hat{x}^{T}-\hat{x}\,\hat{y}%
^{T}\Bigr)+\frac{1}{1+\hat{x}^{T}\hat{y}}\Bigl(\hat{y}\,\hat{x}^{T}-\hat{x}\,%
\hat{y}^{T}\Bigr)^{2}\in \text{O}\left( n\right) ,  \label{R-min}
\end{equation}%
with $\hat{x}:=x/\Vert x\Vert $ and $~\hat{y}:=y/\Vert y\Vert $ denoting the
normalized vectors (\emph{versors}) associated to $x$ resp. $y$, and with $%
I_{n}$ denoting the identity matrix in $n$ dimensions. The \emph{`minimality'%
} of the rotation matrix $\mathcal{\tilde{R}}^{x,y}$ defined by (\ref{R-min}%
) amounts to the fact that it defines a rotation on the plane defined by $%
\hat{x}$ and $~\hat{y}$ \emph{only}. Thus, by setting $\mathcal{R}_{i}=%
\mathcal{\tilde{R}}^{v_{i-1},v_{i}}$, the last vector of the sequence (\ref%
{n-loop}) can be recast into following form ($v^{0}\equiv v$) :
\begin{equation}
v_{L}=\frac{\left\Vert v_{L}\right\Vert }{\left\Vert v\right\Vert }\mathcal{%
\tilde{R}}^{v_{L-1},v_{L}}\mathcal{\tilde{R}}^{v_{L-2},v_{L-1}}...\mathcal{%
\tilde{R}}^{v,v_{1}}v=:\frac{\left\Vert v_{L}\right\Vert }{\left\Vert
v\right\Vert }\mathrm{R}_{\mathcal{U}}^{(\lambda )}v,  \label{thisss}
\end{equation}%
where $\mathrm{R}_{\mathcal{U}}^{(\lambda )}v\in $(S)O$\left( n\right) $
denotes the `minimal'\ rotation matrix associated to the linguistic element $%
\lambda \in \mathscr{A}$ and to the sequence $\mathcal{U}$.

By choosing $d_{\ast }=d_{\ast ,\text{cosine}}$ defined by (\ref{dist}) as
the distance in $\mathbb{R}^{n}$, a little algebra allows to rewrite the
semantic deficit $\delta _{\mathcal{U}}\left( \lambda \right) $ (\ref%
{sem-def}) associated to the linguistic element $\lambda \in \mathscr{A}$
under the action of the sequence $\mathcal{U}$ of linguistic transformations
as a \emph{quadratic form}, denoted by $Q_{\mathcal{U}}\left( \hat{v},\hat{v}%
\right) $ :
\begin{eqnarray}
\delta _{\mathcal{U}}\left( \lambda \right)  &=&Q_{\mathcal{U}}\left( \hat{v}%
,\hat{v}\right) ;  \label{this} \\
Q_{\mathcal{U}}\left( \hat{v},\hat{v}\right)  &:&=\hat{v}^{T}\left( I_{n}-%
\mathrm{R}_{\mathcal{U}}^{(\lambda )\ast }\right) \hat{v},
\end{eqnarray}%
where $\mathrm{R}_{\mathcal{U}}^{(\lambda )\ast }$ is the symmetric part of $%
\mathrm{R}_{\mathcal{U}}^{(\lambda )}$ defined by (\ref{thisss}) :
\begin{equation}
\mathrm{R}_{\mathcal{U}}^{(\lambda )\ast }:=\frac{1}{2}\left( \mathrm{R}_{%
\mathcal{U}}^{(\lambda )}+\mathrm{R}_{\mathcal{U}}^{(\lambda )T}\right) ,
\end{equation}%
and the real symmetric (and not necessarily maximal-rank) matrix $I_{n}-%
\mathrm{R}_{\mathcal{U}}^{(\lambda )\ast }$ is referred to as the \emph{%
representing} matrix of $Q_{\mathcal{U}}\left( \hat{v},\hat{v}\right) $
itself.

The theory of algebraic-geometric invariants of quadratic forms over fields
with characteristic $\neq 2$ is a well developed field of Mathematics \cite%
{Artin,Carmo,Spivak}; \emph{Sylvester's law of inertia} guarantees that the
\emph{signature} sign$\left( I_{n}-\mathrm{R}_{\mathcal{U}}^{(\lambda )\ast
}\right) $ of the matrix $I_{n}-\mathrm{R}_{\mathcal{U}}^{(\lambda )\ast }$
(and thus of the quadratic form $Q_{\mathcal{U}}\left( \hat{v},\hat{v}%
\right) $) is GL$(n,\mathbb{R})$-invariant. By virtue of the \emph{spectral
theorem}, $I_{n}-\mathrm{R}_{\mathcal{U}}^{(\lambda )\ast }$ can be \emph{%
orthogonally diagonalized} : \emph{at least} one (generally, $\mathcal{U}$-
and $\lambda $-dependent) matrix $\mathcal{S}_{\mathcal{U}}^{(\lambda )}\in $%
(S)O$(n)$ exists such that%
\begin{equation}
Q_{\mathcal{U}}\left( \hat{v},\hat{v}\right) =\sum_{a=1}^{n}\rho _{\mathcal{U%
},a}^{(\lambda )}\left( \left( \mathcal{S}_{\mathcal{U}}^{(\lambda )}\hat{v}%
\right) ^{a}\right) ^{2},  \notag
\end{equation}%
where $\rho _{U,a}^{(\lambda )}$'s denote the real eigenvalues of $I_{n}-%
\mathrm{R}_{\mathcal{U}}^{(\lambda )\ast }$. Then, sign$\left( I_{n}-\mathrm{%
R}_{\mathcal{U}}^{(\lambda )\ast }\right) $ is simply provided by the
sequence of the $\rho _{U,a}^{(\lambda )}$'s, in which the non-vanishing
eigenvalues are normalized (but preserve their sign).

The GL$(n,\mathbb{R})$-invariant quantity sign$\left( I_{n}-\mathrm{R}_{%
\mathcal{U}}^{(\lambda )\ast }\right) $ is particularly relevant when the
sequence $\mathcal{U}$ is a linguistic loop : on the one hand, the semantic
deficit $\delta _{\mathcal{U}}\left( \lambda \right) $ measures the overall
`semantic distortion' between the extremal elements $v$ and $v_{L}$ of $\psi
\left( \mathcal{U}\left( \lambda \right) \right) $; on the other hand, the
signature sign$\left( I_{n}-\mathrm{R}_{\mathcal{U}}^{(\lambda )\ast
}\right) $ actually probes finer structural properties of the loop $\mathcal{%
U}$, by the very definition (\ref{thisss}) of the rotation matrix $\mathrm{R}%
_{\mathcal{U}}^{(\lambda )}$. Thus, upon fixing $\mathcal{U}$ itself and
exploiting the invariant sign$\left( I_{n}-\mathrm{R}_{\mathcal{U}%
}^{(\lambda )\ast }\right) $, one can classify in a GL$(n,\mathbb{R})$%
-invariant way the elements of the image $\psi \left( \mathscr{A}\right)
\subsetneq \mathbb{R}^{n}$ of the linguistic space $\mathscr{A}$ under the
embedding map $\psi $.

\section{Conclusions and Future Developments}

In this article, we introduced the concept of linguistic loop in the set of
linguistic transformations of a metrizable linguistic space, and associated
it to an algebraic-geometric invariant in the embedding space : namely, the
signature of the quadratic form expressing the semantic deficit of the loop
itself. This result opens up a totally new line of research, employing tools
from Lie theory and higher-dimensional geometry in the context of linguistic
studies. Our study also introduces a theoretical framework of investigation
of a type of thought at the foundation of many human and animal cognitive
processes, but largely overlooked by contemporary literature : namely, the
\emph{pre-verbal }thought. In our humble opinion, the study of this type of
thought should necessarily be undertaken in order to achieve artificial
intelligences which may \emph{transcend} the usual linguistic manipulation
capabilities, and which may also generalize concepts at a \emph{%
meta-linguistic} level.

The signature associated to the semantic deficit of a linguistic loop reveal
a finer, invariant structure associated, for instance, to two sentences
(namely, the first and the last of the loop) that, despite significant
structural differences, carry virtually identical meaning, as in the case of
translations of the same sentence through a chain of different languages. In
this framework, we put forward the conjecture that the aforementioned
signature captures information on the semantic core of the original (i.e.,
first) sentence, whose modification is controlled and limited throughout the
entire chain of linguistic transformations; as such, the signature should
convey information closely related to the intrinsic, meta-linguistic meaning
of the original sentence itself.

Thus, through the definition of geometric-linguistic invariants and the
concept of semantic deficit associated to each linguistic loop, one could
identify stable conceptual structures, likely to play as key role in a
variety of contexts, from cognitive science, to developmental psychology,
and neuroscience. Remarkably, our approach can be implemented and coded in a
completely explicit way, for instance in the growing efforts to improve the
semantic coherence and robustness of LLM's, especially in contexts that
require complex transformations, such as multilingual translation or
stylistic variation.

Consequently, it is easy to realize that the invariant analysis of the
semantic deficit associated to a linguistic loop, performed in terms of
algebraic-geometric tools pertaining to Lie theory and higher-dimensional
geometry, provides a basis for developing new metrics aimed at evaluating
the robustness of linguistic transformations, as well as at creating latent
representations that could constitute the `seeds'{} of an \emph{abstract}
thought; intriguingly, this might provide a theoretical bridge between the
non-verbal thought and its linguistic expression. It is also here worth
pointing out that the use of these concepts in the context of model
distillation would provide practical tools for the compression and
optimization of LLM's themselves. On the other hand, we should also remark
that our approach, due to its theoretical and foundational nature, still
requires a large-scale empirical validation, in which its fertility and
relevance may be concretely explored and investigated..

All in all, we have presented evidence that the integration of linguistic
transformations with invariant analysis from Lie theory and
higher-dimensional geometry may potentially bridge the notoriously difficult
gap between the pre-verbal thought and the linguistic expression. The next
steps of our investigation will clearly include the empirical verification
of these concepts, as well as the exploration of some of their practical
implications, with the aim of expanding our tools for the understanding of
human thought and artificial intelligence.

\end{document}